\documentclass[letterpaper]{article} 
\usepackage[draft]{aaai2027}
\usepackage[hyphens]{url}  
\usepackage{graphicx} 
\usepackage{natbib}  
\usepackage{caption} 
\usepackage{algorithm}

\usepackage{amsmath,amssymb,amsfonts}
\usepackage{algpseudocode}  
\usepackage{multirow}
\usepackage{enumitem}

\usepackage{newfloat}
\usepackage{listings}
\DeclareCaptionStyle{ruled}{labelfont=normalfont,labelsep=colon,strut=off} 
\floatstyle{ruled}
\newfloat{listing}{tb}{lst}{}
\floatname{listing}{Listing}

\usepackage{booktabs}

\title{Move What Matters: Parameter-Efficient Domain Adaptation via Optimal Transport Flow for Collaborative Perception}
\author{
    Zesheng Jia\textsuperscript{1},
    Jin Wang\textsuperscript{1}\thanks{Corresponding authors.},
    Siao Liu\textsuperscript{1}\footnotemark[1],
    Lingzhi Li\textsuperscript{1},
    Ziyao Huang\textsuperscript{2},
    Yunjiang Xu\textsuperscript{2},
    Jianping Wang\textsuperscript{2}
}
\affiliations{
    \textsuperscript{\rm 1}School of Future Science and Engineering, Soochow University\\
    \textsuperscript{\rm 2}City University of Hong Kong\\
    \{wjin1985, saliu\}@suda.edu.cn

}

\begin{document}

\maketitle

\begin{abstract}
Efficient domain adaptation remains a fundamental challenge for deploying multi-agent systems across diverse environments in Vehicle-to-Everything (V2X) collaborative perception. Despite the success of Parameter-Efficient Fine-Tuning (PEFT) in natural language processing and conventional vision tasks, directly applying PEFT to collaborative perception recovers only a limited portion of the performance lost to domain shift. In this work, we identify two complementary bottlenecks that limit this recovery: (i) inter-frame redundancy within a collaborative sequence, which makes the effectiveness of scarce labels sensitive to frame selection, and (ii) foreground cues that become less linearly decodable in deeper-stage representations of a frozen backbone. To address these issues, we propose FlowAdapt, a parameter-efficient framework grounded in optimal transport. Wasserstein Greedy Sampling casts frame selection as minimizing the $W_\infty$ distance from the sequence to the retained subset, which equals its covering radius, so a farthest first traversal returns a subset provably within twice the optimum. Progressive Knowledge Transfer then routes compressed early-stage features into the deeper stages, gating each stage-local correction by that early evidence. Extensive experiments across target domains and fusion architectures show that FlowAdapt achieves state-of-the-art adaptation performance with about 1\% trainable parameters, and maintains this lead under localization noise.
\end{abstract}

\section{Introduction}
\label{sec:intro}

Autonomous vehicles suffer from inherent perceptual limitations including occlusions, restricted field-of-view, and limited sensing range~\cite{huangV2XCooperativePerception2024}. Vehicle-to-Everything (V2X) collaborative perception bridges the gap between isolated single-agent observation~\cite{caillotSurveyCooperativePerception2022} and comprehensive environmental awareness through multi-agent information sharing~\cite{chenFcooperFeatureBased2019}. However, most such detectors are trained and deployed under the tacit assumption that both stages share one sensor setup and one band of conditions, which deployment rarely honors~\cite{songTraFalignTrajectoryawareFeature2025}. A new city or a different sensor shifts the input distribution away from the one the frozen features were fitted to, and detection accuracy drops sharply~\cite{weiCoPEFTFastAdaptation2025}. Retraining is the obvious recourse, yet it demands annotations the new domain has yet to receive and a separate model per domain thereafter. Cross-domain adaptation cheap in both respects remains an open problem for collaborative perception.

\begin{figure}[t]
\centering
\includegraphics[width=\linewidth]{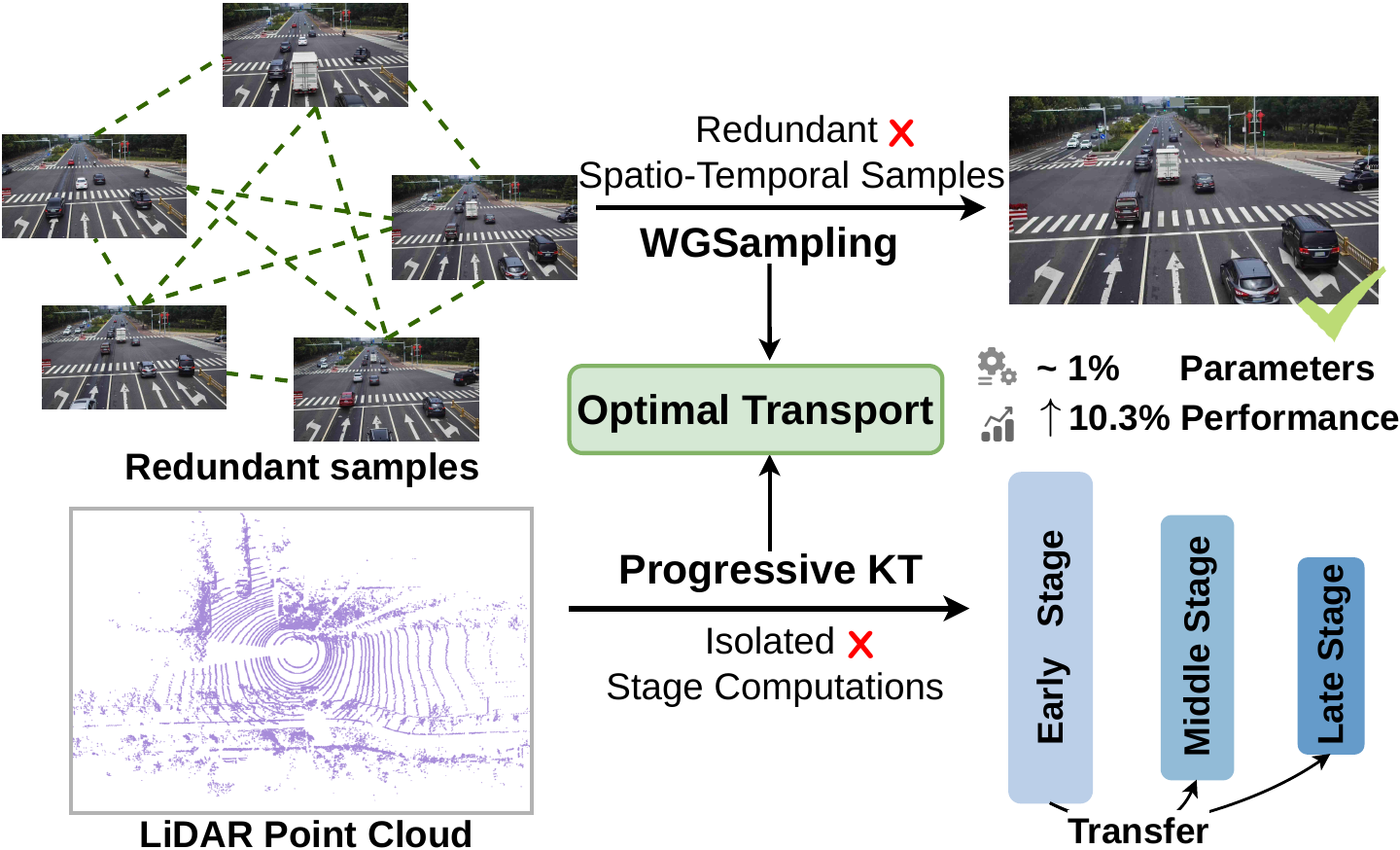}
\caption{FlowAdapt answers two obstacles to parameter-efficient adaptation. Wasserstein Greedy Sampling chooses which target frames to label, and Progressive Knowledge Transfer routes the early features to the deeper stages.}
\vspace{-5pt}
\label{fig:small_image}
\end{figure}

\begin{figure}[t]
\centering
\includegraphics[width=\linewidth]{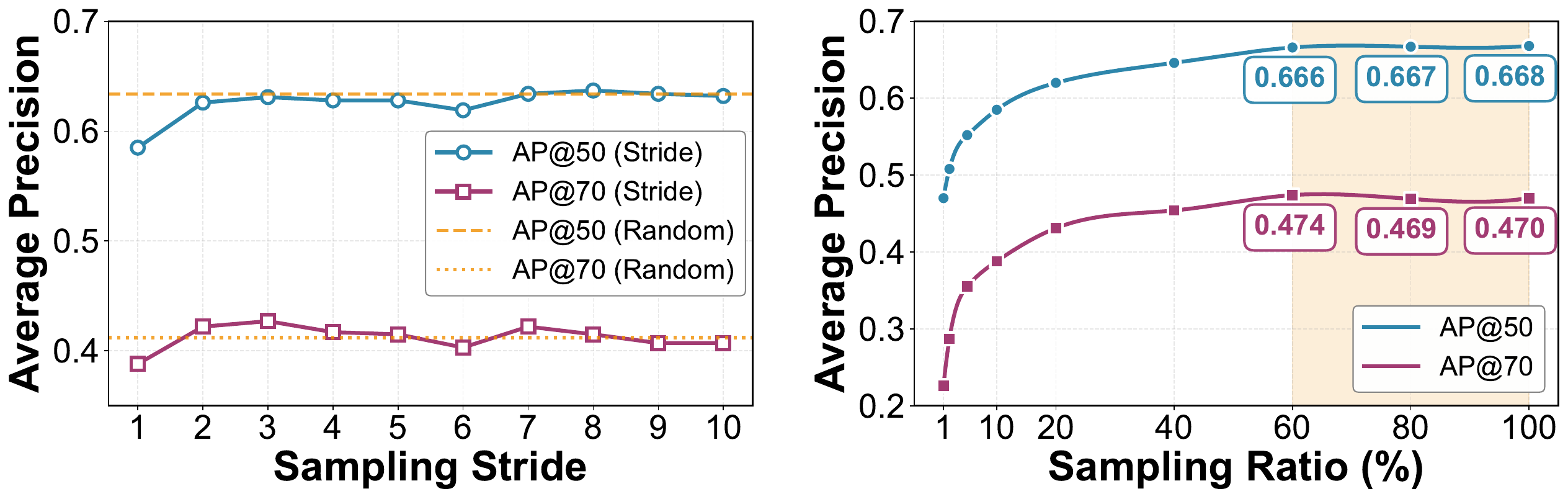}
\caption{Left: performance versus temporal stride, the interval between consecutively selected frames, at a fixed 10\% sampling ratio. Right: performance versus sampling ratio under sequential selection, flattening beyond 60\%.}
\label{fig:visual_wgs}
\end{figure}

\begin{figure}[t]
\centering
\includegraphics[width=\linewidth]{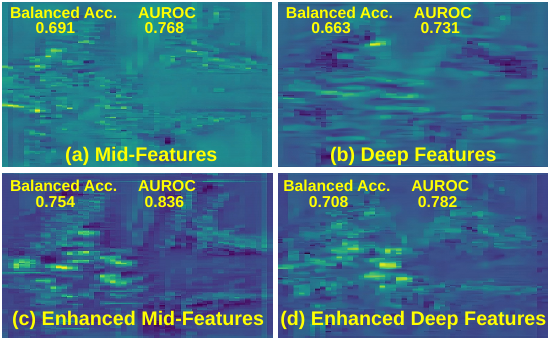}
\caption{Feature responses at two depths of the frozen backbone during adaptation, middle (within the backbone) and deep (before the detection head). A linear probe is trained at each depth to predict from every cell whether it contains a foreground object, scored by balanced accuracy and AUROC. The probe degrades with depth (a, b) and recovers once compressed early features are injected (c, d), so decodability at depth is open to a cross-stage intervention.}
\label{fig:visual_kdpro}
\end{figure}

To overcome this limitation, parameter-efficient fine-tuning (PEFT) adapts a pre-trained model while keeping most of it frozen, training only a small set of parameters. Adapters~\cite{adapter_houlsby} and LoRA~\cite{hu2021loralowrankadaptationlarge} established the approach on language models and were subsequently adopted in vision~\cite{xin2025parameterefficientfinetuningpretrainedvision,he2023sensitivityawarevisualparameterefficientfinetuning}. MACP~\cite{maMACPEfficientModel2024} attaches lightweight modules to a frozen single-agent detector so that it can collaborate, while CoPEFT~\cite{weiCoPEFTFastAdaptation2025} adapts a trained collaborative detector to a new domain through adapters and prompts placed across the fusion hierarchy. Examining CoPEFT, the strongest PEFT baseline in our setting, at a small labeling budget, we make two complementary observations. (i) \textbf{\textit{Inter-frame redundancy}} within a collaborative sequence. As illustrated in Figure~\ref{fig:visual_wgs}, accuracy rises steeply over the first few percent of labeled frames and then levels off, while holding the ratio fixed and varying the stride alone moves AP@50 by 5.2 points and AP@70 by 3.9 points. (ii) \textbf{\textit{Fading foreground decodability}} in the deep layers. With the backbone frozen, a linear probe reads foreground evidence off the deeper representations far less reliably than off the shallower ones (Figure~\ref{fig:visual_kdpro}), so a downstream adapter is left to work on features whose foreground cues are harder to recover by a linear map. Combining those features with a compressed form of the early ones lifts the probe by four to seven points in balanced accuracy, so decodability at depth is open to intervention and a cross-stage route is one way to intervene.

In this work, we view adaptation as a matter of transport under two distinct scarcities, of labels and of parameters. The first observation is a transport problem in the literal sense: a subset stands in for a sequence to the degree that no frame lies far from the one representing it, and that worst displacement is the smallest $\infty$-Wasserstein distance from the empirical measure of the sequence to any measure supported on the subset, so selecting frames amounts to minimizing it over the subsets of a given size. The second instead concerns where the remaining parameters go. Since foreground cues grow harder to read out along the frozen hierarchy, part of the budget buys a route from the early stages to the deeper ones, whose local corrections it modulates rather than replaces.

Based on these insights, we propose FlowAdapt, a parameter-efficient framework whose two components answer the two observations in turn. Wasserstein Greedy Sampling (WGS) reads a frame as a collaborative configuration, describing when it was taken, how many agents were present, how they were arranged and what they jointly observed, and seeks the subset minimizing that transport distance. Since the distance equals the covering radius of a subset, turning a transport objective into a covering one, a farthest-first traversal returns a subset whose radius is within twice the optimum, whatever the starting frame. Progressive Knowledge Transfer (KTPro) opens the route the second observation asks for: the adapted early features are cached as a detached memory, which a compressor at each deeper stage pools and narrows, and an injector there expands and resamples into a gate on the correction that stage's adapter proposes, leaving the frozen features it joins untouched. Across three target domains and three fusion architectures, FlowAdapt reaches state-of-the-art adaptation accuracy with only 1\% trainable parameters. In summary, our contributions are threefold:

\begin{itemize}[leftmargin=*]
    \item We identify two obstacles to parameter-efficient adaptation in collaborative perception. \textbf{\textit{Inter-frame redundancy}}: neighboring frames overlap heavily, and two selections of equal size differ by several points of accuracy, so which frames are annotated is part of the problem rather than a given. \textbf{\textit{Fading foreground decodability}}: a linear probe recovers foreground cues from the deeper stages of a frozen backbone far less reliably than from the shallower ones.
    \item We propose FlowAdapt, which addresses both under an optimal transport view. WGS casts selection as a covering problem over collaborative configurations, for which a greedy traversal returns a solution within a factor of two of the optimum. KTPro compresses the adapted early features into a gate on the correction each deeper adapter proposes.
    \item We evaluate FlowAdapt across three target domains and three fusion architectures, and show that it reaches state-of-the-art adaptation accuracy with 1\% trainable parameters while holding its lead under localization noise.
\end{itemize}

\section{Related Work}

\subsection{Collaborative Perception}
Collaborative perception lets vehicles and roadside units share what a single viewpoint cannot observe~\cite{chenFcooperFeatureBased2019,liu2023vehicletoeverythingautonomousdrivingsurvey}. What they exchange sets the trade-off. Raw returns preserve detail at the cost of bandwidth~\cite{cooper_chen_early}, detections travel cheaply but discard the evidence behind them~\cite{car2x_late}, and intermediate features balance the two~\cite{wangV2VNetVehicletovehicleCommunication2020,xu2022cobevtcooperativebirdseye}. Their exchange, however, invites much that can go wrong. Bandwidth is budgeted by choosing whom to contact and over which regions~\cite{liuWho2comCollaborativePerception2020,huWhere2commCommunicationefficientCollaborative2022}, or by sending instance-level queries in place of dense maps~\cite{xu2025instinct}. Misregistered messages are realigned against pose error through agent-object graphs~\cite{luRobustCollaborative3D2023}, and latency is met by warping features along trajectories~\cite{songTraFalignTrajectoryawareFeature2025} or by attention built to tolerate delay and pose error together~\cite{xuV2XViTVehicletoeverythingCooperative2022}. HEAL closes a domain gap of a different kind, admitting agents of unlike sensors and backbones by aligning each newcomer to a space the collaboration already shares~\cite{lu2024extensibleframeworkopenheterogeneous}, where our concern is a collaboration whose scene and sensing have both moved, and what a small budget of labels buys there.

\subsection{Parameter-Efficient Fine-Tuning}
Parameter-efficient fine-tuning adapts a largely frozen pre-trained model by training only a small number of parameters~\cite{lialin2024scalingscaleupguide}. Adapters insert bottleneck layers inside each transformer block~\cite{adapter_houlsby}, LoRA and its variants reparameterize the weight update~\cite{hu2021loralowrankadaptationlarge,shi2024reslora}, and prefix tuning learns key and value prefixes within every attention layer~\cite{li2021prefixtuningoptimizingcontinuousprompts}, each approaching full fine-tuning in its own domain, vision among them~\cite{jia2022visualprompttuning,adapter_chen}. Ladder side-tuning instead runs a separate branch alongside the frozen network, predicting from activations it reads off, so that no gradient need traverse the backbone~\cite{sung2022lst}. In collaborative perception, MACP grafts modules onto a frozen single-agent detector to let it collaborate~\cite{maMACPEfficientModel2024}, while CoPEFT, the closest work to ours, adapts a trained collaborative detector to a new domain through a Collaboration Adapter and an Agent Prompt placed across the fusion hierarchy~\cite{weiCoPEFTFastAdaptation2025}. Neither asks which target frames are worth annotating, nor how linearly decodable the features are that its adapters receive.

\subsection{Optimal Transport}
Optimal transport compares two distributions by the cost of moving one onto the other, requiring no known correspondence between their samples~\cite{peyr2020computationaloptimaltransport,khamis2024scalable}. It reached domain adaptation by transporting source samples onto the target under regularizers that respect label structure~\cite{courty2016optimaltransportdomainadaptation,courty2017jointdistributionoptimaltransportation}, and entropic regularization of the transport objective brought large instances within reach~\cite{cuturi2013sinkhorn}. Transport cost also serves to choose and compare training data. GORACS scores a coreset by a proxy for test loss built from transport cost and gradient information~\cite{mei2025goracsgroupleveloptimaltransportguided}, while Multi-Level OT distils between models whose tokenizers do not align~\cite{cui2025multileveloptimaltransportuniversal}. Both rest on a finite-order Wasserstein distance, which permits a trade, a few configurations left far away in exchange for many placed close. Under a small annotation budget the worst case admits no such trade. We formulate frame selection as minimizing the smallest $\infty$-Wasserstein distance from the empirical measure of a sequence to any measure supported on the frames actually retained.

\section{Methodology}

\begin{figure*}[t]
\centering
\includegraphics[width=\textwidth]{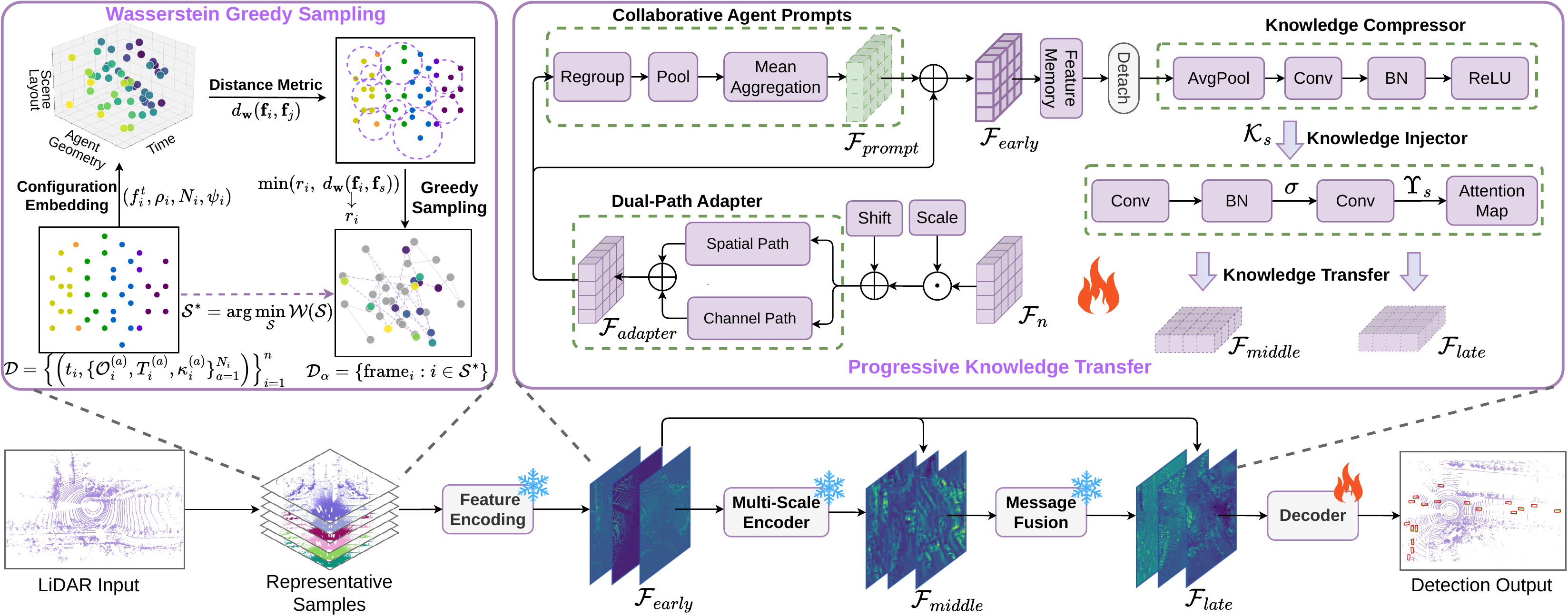}
\caption{
Overview of FlowAdapt, a parameter-efficient framework that adapts a collaborative detector under a small annotation budget while its source representation stays frozen. Wasserstein Greedy Sampling decides which target frames to label, casting the choice as a transport problem over collaborative configurations. Progressive Knowledge Transfer opens a compressed route from the early stage to the deeper ones, where it gates the corrections that stage-local adapters propose.
}
\label{fig:overview}
\end{figure*}

\subsection{Overall Architecture}
\label{sec:overall}
In collaborative perception for V2X systems, a set of $N$ agents $\mathcal{A}=\{A_{1},\dots,A_{N}\}$ is present in the scene, each holding a LiDAR observation $\mathcal{O}_{j}$ and a pose $T_{j}$. The goal is to improve 3D object detection for a designated ego $A_{i}$ through the cooperative sharing of complementary information among agents, and this paper focuses on intermediate collaboration. Our concern is to adapt such a detector from a source domain to a target one under a small annotation budget, leaving the source representation frozen. Figure~\ref{fig:overview} gives an overview, and the pipeline proceeds as
\begin{subequations}
\label{eq:pipeline}
\begin{align}
\mathcal{D}_{\alpha} &= \mathrm{WGS}(\mathcal{D},\alpha), \label{eq:pipeline_a}\\
\mathcal{F}^{e}_{j} &= f_{\text{enc}}(\mathcal{O}_{j}), \quad
\widehat{\mathcal{F}}^{e}_{j} = f^{\text{early}}_{\text{adapt}}\big(\mathcal{F}^{e}_{j},\,\{\mathcal{F}^{e}_{k}\}_{k\in g}\big), \label{eq:pipeline_b}\\
\mathcal{M}^{e}_{j} &= \mathrm{detach}\big(\widehat{\mathcal{F}}^{e}_{j}\big), \quad
\mathcal{F}^{m}_{j} = f_{\text{bb}}\big(\widehat{\mathcal{F}}^{e}_{j}\big), \label{eq:pipeline_c}\\
\widehat{\mathcal{F}}^{m}_{j} &= f^{\text{mid}}_{\text{adapt}}\big(\mathcal{F}^{m}_{j},\, \mathcal{M}^{e}_{j}\big), \quad
\mathcal{F}_{j\to i} = \Gamma_{j\to i}\big(\widehat{\mathcal{F}}^{m}_{j}\big), \label{eq:pipeline_d}\\
\mathcal{H}_{i} &= f_{\text{fus}}\big(\{\mathcal{F}_{j\to i}\}_{j=1}^{N}\big), \quad
\widehat{\mathcal{H}}_{i} = f^{\text{late}}_{\text{adapt}}\big(\mathcal{H}_{i},\, \mathcal{M}^{e}_{i}\big), \label{eq:pipeline_e}\\
\widehat{\mathcal{B}}_{i} &= f_{\text{det}}\big(\widehat{\mathcal{H}}_{i}\big). \label{eq:pipeline_f}
\end{align}
\end{subequations}
WGS selects from the target sequence $\mathcal{D}$ a subset $\mathcal{D}_{\alpha}$ of $\lfloor\alpha|\mathcal{D}|\rfloor$ frames, on which all training below is performed. The frozen encoder $f_{\text{enc}}$ turns a point cloud into bird's eye view features, and $f^{\text{early}}_{\text{adapt}}$ adapts them under a prompt pooled from $g$, the agents sharing the frame. Its output continues along the main path and is also held as $\mathcal{M}^{e}_{j}$, the same values detached, so gradients reach the early adapter through the main path only. The frozen backbone $f_{\text{bb}}$ abstracts these features further, and $f^{\text{mid}}_{\text{adapt}}$ adapts the result while injecting the memory through a compressor and an injector of its own. Each agent's features are then resampled into the ego frame by $\Gamma_{j\to i}$, the warping induced by the relative pose $T_{i}^{-1}T_{j}$, and combined by the frozen fusion $f_{\text{fus}}$ into a collaborative representation $\mathcal{H}_{i}$. A second injection follows at $f^{\text{late}}_{\text{adapt}}$, which reads $\mathcal{M}^{e}_{i}$ alone, since $\mathcal{H}_{i}$ and the ego memory share that frame while the other memories do not. The detection head $f_{\text{det}}$ produces $\widehat{\mathcal{B}}_{i}$ and is trained alongside the adapters.


\subsection{Wasserstein Greedy Sampling}
Frames in a collaborative sequence are far from equally informative. Over short spans the scene evolves slowly and the geometry among agents shifts little, so nearby frames contribute overlapping supervision. Adaptation is better served by a compact subset in which every sample stays close to a representative, and Wasserstein Greedy Sampling selects one under a transport cost over collaborative configurations.

\subsubsection{A Ground Metric on Collaborative Configurations.}
A frame in the multi-agent setting is characterized by when it was captured, how the agents were arranged, and what they jointly observed. The arrangement admits no natural vectorization, the number of agents varying across frames and their ordering carrying no meaning, so we describe it by a measure over the agents, compared by a transport cost. Let $N_i$ agents be present in frame $i$, with $T^{(a)}_i$ the pose of agent $a$ and $\kappa^{(a)}_i\in\{0,1\}$ marking it as vehicle or infrastructure. Writing the relative pose $(T^{\mathrm{ego}}_i)^{-1}T^{(a)}_i$ as a planar offset $(u^{(a)}_i,v^{(a)}_i)$ and a heading $\theta^{(a)}_i$, we embed the agent as
\begin{equation}
q^{(a)}_i=\big[\tfrac{u^{(a)}_i}{L},\ \tfrac{v^{(a)}_i}{L},\ \tfrac{\cos\theta^{(a)}_i}{2},\ \tfrac{\sin\theta^{(a)}_i}{2},\ \tau\kappa^{(a)}_i\big]^{\!\top}\in\mathcal{Q},
\label{eq:agent_embed}
\end{equation}
with $L$ the communication range. Heading enters through the circle, so orientations near $\pm\pi$ are seen as close, and $\tau$ exceeds $\sqrt5$, the diameter of the first four coordinates, which prices transport across types above every displacement within one while keeping type and pose in a single descriptor. The image $\mathcal{Q}$ is a product of a unit disc, a circle of radius $\tfrac12$ and a two point set, compact with $\operatorname{diam}(\mathcal{Q})=\sqrt{5+\tau^{2}}$ following from its factors. A collaboration is the uniform measure $\rho_i=\frac{1}{N_i}\sum_{a}\delta_{q^{(a)}_i}$ on $\mathcal{Q}$, and two are compared by $W_2(\rho_i,\rho_j)$ under the Euclidean ground metric, which weighs them even when they hold unequally many agents.

Unit mass leaves $\rho_i$ a description of arrangement alone, so the count $N_i$ is kept apart from it. Geometry is in turn blind to the scene, which we capture by warping the observations $\{\mathcal{O}^{(a)}_i\}$ into the ego frame and rasterizing their union onto a bird's eye view grid, marking every occupied cell. The resulting $\psi_i\in\{0,1\}^{HW}$ distinguishes frames observed from identical poses but differing in what the sensors returned. With a timestamp $f^t_i$ normalized over the sequence, a configuration reads $\mathbf{f}_i=(f^t_i,\rho_i,N_i,\psi_i)$, compared by
\begin{equation}
d_{\mathbf{w}}(\mathbf{f}_i,\mathbf{f}_j)=w_t\,\delta^{t}_{ij}+w_g\,\delta^{g}_{ij}+w_n\,\delta^{n}_{ij}+w_c\,\delta^{c}_{ij}.
\label{eq:ground_metric}
\end{equation}
WGS runs within each sequence, drawing a share of the budget proportional to its length, and \eqref{eq:ground_metric} compares the configurations of one such sequence. The four discrepancies read timestamps, arrangements, counts and scenes, $\delta^{t}_{ij}=|f^t_i-f^t_j|$, $\delta^{g}_{ij}=W_2(\rho_i,\rho_j)/\operatorname{diam}(\mathcal{Q})$, $\delta^{n}_{ij}=|N_i-N_j|/N_{\max}$ with $N_{\max}$ the largest count it holds, and $\delta^{c}_{ij}=\lVert\psi_i-\psi_j\rVert_1/HW$, each scaled to $[0,1]$ so that the strictly positive $\mathbf{w}$, fixed across domains along with $\tau$, governs influence rather than units. Each is a metric on its factor, and a positive combination of metrics is a metric, so $d_{\mathbf{w}}$ makes the configurations of a sequence a metric space $\mathcal{F}$.

\subsubsection{Selection as Optimal Transport.}
Retaining a subset replaces the sequence by a measure supported on the frames kept. Write $\mu_n=\frac1n\sum_{i}\delta_{\mathbf{f}_i}$ for the empirical measure over $\mathcal{F}$, let $\mathcal{S}$ of size $m$ index the retained frames, and write $\mathcal{F}_{\mathcal{S}}=\{\mathbf{f}_j:j\in\mathcal{S}\}$ for the configurations they carry. Since no configuration may be left unrepresented, a surrogate is judged by its largest displacement rather than its average,
\begin{equation}
W_\infty(\mu_n,\nu)=\inf_{\gamma\in\Pi(\mu_n,\nu)}\ \sup_{(\mathbf{f},\mathbf{f}')\in\operatorname{supp}\gamma}d_{\mathbf{w}}(\mathbf{f},\mathbf{f}'),
\label{eq:winf}
\end{equation}
with $\Pi$ the couplings matching the marginals, and the subset itself by the best surrogate it admits, $\mathcal{W}(\mathcal{S})=\inf_{\operatorname{supp}\nu\subseteq\mathcal{F}_{\mathcal{S}}}W_\infty(\mu_n,\nu)$. Sending every frame to its nearest retained one costs no more than the covering radius $R(\mathcal{S})=\max_i\min_{j\in\mathcal{S}}d_{\mathbf{w}}(\mathbf{f}_i,\mathbf{f}_j)$, while every frame carries mass $1/n>0$ that no coupling can move for less than its own distance to $\mathcal{F}_{\mathcal{S}}$. The two bounds meet, so that
\begin{equation}
\mathcal{W}(\mathcal{S})\;=\;R(\mathcal{S}),
\label{eq:equiv}
\end{equation}
which turns selection into
\begin{equation}
\mathcal{S}^{*}=\arg\min_{|\mathcal{S}|=m}\ \mathcal{W}(\mathcal{S})=\arg\min_{|\mathcal{S}|=m}\ R(\mathcal{S}),
\label{eq:wgs_obj}
\end{equation}
an $m$-center problem on the metric space $(\mathcal{F},d_{\mathbf{w}})$. Since $W_p\le W_\infty$ for every finite $p$, the choice bounds the whole family, whereas minimizing a finite $p$ would tolerate a handful of frames left far from any representative, a small mass over a long distance weighing little in an averaged cost.

\subsubsection{Greedy Selection.}
Problem \eqref{eq:wgs_obj} is NP-hard over general metrics~\cite{kleindessner2019fair}, and WGS approaches it by farthest-first traversal, starting from any frame and repeatedly retaining the configuration worst covered by those already kept, then refreshing the radii. The classical guarantee on the covering radius, carried over to transport by \eqref{eq:equiv}, places the result within twice the optimum,
\begin{equation}
\mathcal{W}\big(\mathcal{S}_{\mathrm{WGS}}\big)\ \le\ 2\min_{|\mathcal{S}|=m}\ \mathcal{W}(\mathcal{S}).
\label{eq:approx}
\end{equation}
Radii refreshed incrementally, the pairwise distance matrix is never formed, and selection costs $O(nm)$ evaluations of \eqref{eq:ground_metric} in $O(n)$ working memory. Proof is in the \textbf{Appendix}.


\subsection{Progressive Knowledge Transfer}
A linear probe decodes foreground cues from the deeper stages of the frozen backbone far less reliably than from the shallower ones (Figure~\ref{fig:visual_kdpro}), and the adapters placed there must propose corrections on features whose foreground structure is harder to read out. Progressive Knowledge Transfer (KTPro) spends a small parameter budget on a compressed side route, which reads the adapted early features and gates, at each deeper stage, the correction its adapter proposes.

\subsubsection{Dual-Path Adapter.}
A change of domain moves evidence over the bird's eye view and reshapes how the channels respond, and the two need not be separable. We nonetheless treat them apart, giving a tight bottleneck one operator for each rather than a single pathway serving both. The adapter at stage $\ell$ projects the width $C$ down to $C/r_{\ell}$ through a single $\mathrm{Down}$, sends that projection along two branches, and restores the width as a residual,
\begin{equation}
\mathcal{G}(\mathcal{F})=\mathrm{Up}\Big(\lambda_{s}\,\mathrm{S}\big(\mathrm{Down}(\mathcal{F})\big)+\lambda_{c}\,\mathrm{C}\big(\mathrm{Down}(\mathcal{F})\big)\Big),
\label{eq:dual_path_fusion}
\end{equation}
where $\mathrm{S}$ stacks grouped convolutions, acting along space and mixing channels only within a group, and $\mathrm{C}$ stacks pointwise projections, recombining channels at every position without mixing across them. The weights $(\lambda_{s},\lambda_{c})$ are a softmax over two scalars learned per adapter. This $\mathcal{G}$ is the correction the stage proposes, added to the frozen features in \eqref{eq:gate} and scaled by the route of Section~\ref{sec:cross_stage}. Capacity is not spread evenly. The early stage carries $N_{\text{early}}=3$ blocks at the least compressed bottleneck, adapting the earliest representation the encoder produces, while the middle and late stages carry one each at $r_{\text{early}}<r_{\text{mid}}<r_{\text{late}}$, their inputs already adapted upstream.

\subsubsection{Cross-Stage Transfer.}
\label{sec:cross_stage}
The route carries the detached memory $\mathcal{M}^{e}$ to the middle and late stages, each with a compressor and an injector of its own. Crossing a change of resolution and of width, the \textit{compressor} pools the memory onto an $8\times20$ grid and contracts the channels by $r_{c}$,
\begin{equation}
\mathcal{K}_{s}=\mathrm{ReLU}\big(\mathrm{BN}(\mathrm{Conv}(\mathrm{AdaptiveAvgPool}(\mathcal{M}^{e})))\big),
\label{eq:compress}
\end{equation}
so that structure at the scale of the grid is what survives, the middle stage reading each agent's memory and the late stage the ego's alone, as \eqref{eq:pipeline} prescribes. The \textit{injector} expands $\mathcal{K}_{s}$ to the width of stage $s$ and resamples it to that resolution,
\begin{equation}
\mathcal{A}_{s}=\Upsilon_{s}\Big(\sigma\big(\mathrm{Conv}(\mathrm{ReLU}(\mathrm{BN}(\mathrm{Conv}(\mathcal{K}_{s}))))\big)\Big),
\label{eq:inject}
\end{equation}
with $\sigma$ the sigmoid, placing $\mathcal{A}_{s}$ entrywise in $(0,1)$, and $\Upsilon_{s}$ bilinear resampling. The result gates the correction $\mathcal{G}_{s}$ before it rejoins the frozen path,
\begin{equation}
\widehat{\mathcal{F}}_{s}=\mathcal{F}_{s}+\beta_{s}\,\big[\mathcal{G}_{s}\odot(1+\alpha_{s}\mathcal{A}_{s})\big],
\label{eq:gate}
\end{equation}
with $\alpha_{s}$ and $\beta_{s}$ scalars clamped to $[0.1,0.5]$. The gate factor lies in $(1,1.5)$, so the memory scales the correction entrywise without changing the sign of an entry, and what the adapter leaves at zero it cannot move. The route amplifies a correction where the memory responds most strongly rather than proposing its own. That route runs one way. The memory enters detached, so the injectors open no second route back to the early adapter, which would let the deeper stages pull the early features toward whatever eased their fitting, though those features are also what the frozen backbone consumes.

\subsubsection{Collaborative Agent Prompts.}
The early adapter treats one agent at a time, so we condition it on a statistic of its group rather than a prompt learned free of the data. With $\bar{\mathcal{F}}^{e}_{j}$ its output before conditioning and $g$ the agents of the frame, each is pooled over the bird's eye view into a channel response, whose projected mean is broadcast over the grid,
\begin{equation}
\widehat{\mathcal{F}}^{\,e}_{j}=\bar{\mathcal{F}}^{\,e}_{j}+\gamma\,\mathrm{Bcast}\Big(\phi\Big(\tfrac{1}{|g|}\textstyle\sum_{k\in g}\mathrm{Pool}\big(\bar{\mathcal{F}}^{\,e}_{k}\big)\Big)\Big),
\label{eq:agent_prompts}
\end{equation}
with $\phi$ a lightweight projection and $\gamma=0.1$. Pooling before the average spares the prompt any registration, so what crosses the group is a summary of how its members respond, unchanged by their ordering and carrying no trace of where each found its evidence. Entering as a residual, the prompt shifts a member's channels and leaves the rest to it.

\section{Experiments}

\subsection{Implementation Details}

\subsubsection{Datasets}
Models are source-trained on OPV2V~\cite{xu2022opv2vopenbenchmarkdataset}, a vehicle-to-vehicle simulation built with CARLA~\cite{dosovitskiy2017carlaopenurbandriving} and OpenCDA~\cite{xu2021opencdaanopencooperativedriving}, 11K frames of two to seven agents, then adapted to three targets whose shifts differ in kind. V2XSet~\cite{xuV2XViTVehicletoeverythingCooperative2022} stays in simulation but adds roadside infrastructure, making the collaboration heterogeneous. DAIR-V2X~\cite{dairv2x} is real and vehicle to infrastructure, one vehicle with one roadside unit under sensor noise and temporal asynchrony. V2V4Real~\cite{v2v4real} is real and vehicle to vehicle, two cars over 410\,km of highway and city roads. The three thus span a simulated shift in collaboration structure and two real ones differing in who collaborates. For DAIR-V2X we follow the protocol of~\cite{luRobustCollaborative3D2023} with the supplementary annotations of~\cite{xiang2023div2xlearningdomaininvariantrepresentation}.

\subsubsection{Evaluation Metrics.}
We report Average Precision on the car class at IoU thresholds of 0.5 and 0.7 (AP@50, AP@70), evaluated in the ego frame over $x\in[-100,100]$\,m and $y\in[-40,40]$\,m. To probe robustness under localization uncertainty, we perturb each non-ego agent's pose independently at inference, adding zero-mean Gaussian noise of standard deviation $\sigma_p$ to its planar translation and $\sigma_y$ to its yaw before the ego-frame transformation is rebuilt.

\subsubsection{Training Configuration.}
Source models are trained on OPV2V with CoAlign~\cite{luRobustCollaborative3D2023}, unless another fusion architecture is under study, on point clouds voxelized into pillars of $0.4\times0.4$\,m. Adaptation runs from OPV2V to DAIR-V2X with 10\% of the target samples labeled, unless stated otherwise, at a batch size of four on two NVIDIA RTX 4080 GPUs. Each deeper stage owns a compressor and an injector, while the early stage carries $N_{\text{early}}=3$ adaptation blocks and the middle and late stages one each, at bottleneck reduction ratios $r_{\text{early}}<r_{\text{mid}}<r_{\text{late}}$. We update only the adapter modules and the decoder. For reference, the unsupervised baseline DUSA~\cite{kongDUSADecoupledUnsupervised2023} requires the whole target pool unlabeled to reach competitive performance.

\begin{table*}[t]
\centering
{
\small
\setlength{\tabcolsep}{8pt}
\renewcommand{\arraystretch}{0.95}
\begin{tabular}{@{}lcccccc@{}}
\toprule
Method
  & 1\%
  & 2\%
  & 5\%
  & 10\%
  & 20\%
  & Parameter \\
\cmidrule(lr){2-6}
  & \multicolumn{5}{c}{AP@50/70~$\uparrow$} & \\
\midrule

None
  & 0.429/0.218
  & 0.429/0.218
  & 0.429/0.218
  & 0.429/0.218
  & 0.429/0.218
  & 0/12,896,384 = 0.00\% \\

From scratch
  & 0.141/0.054
  & 0.201/0.070
  & 0.336/0.138
  & 0.427/0.212
  & 0.591/0.397
  & 12,896,384/12,896,384 = 100.00\% \\

Decoder only
  & 0.431/0.191
  & 0.457/0.229
  & 0.484/0.235
  & 0.482/0.256
  & 0.519/0.286
  & 5,140/12,901,524 = 0.04\% \\

SSF
  & 0.509/0.231
  & 0.508/0.263
  & 0.523/0.276
  & 0.537/0.298
  & 0.561/0.329
  & 5,780/12,902,164 = 0.04\% \\

Adapter
  & 0.484/0.229
  & 0.495/0.260
  & 0.534/0.282
  & 0.550/0.322
  & 0.593/0.366
  & 42,420/12,938,804 = 0.33\% \\

LoRA
  & 0.239/0.064
  & 0.321/0.142
  & 0.453/0.245
  & 0.545/0.346
  & 0.609/0.441
  & 457,748/13,354,132 = 3.43\% \\

ResLoRA
  & 0.254/0.073
  & 0.327/0.144
  & 0.467/0.255
  & 0.557/0.355
  & 0.625/0.457
  & 684,052/13,580,436 = 5.04\% \\

DUSA
  & -
  & -
  & -
  & 0.530/0.314
  & -
  & 14,213,266/14,213,266 = 100.00\% \\

MACP
  & 0.479/0.251
  & 0.506/0.289
  & 0.556/0.337
  & 0.577/0.382
  & 0.620/0.412
  & 43,060/12,939,444 = 0.33\% \\

CoPEFT
  & 0.505/0.256
  & 0.521/0.298
  & 0.586/0.357
  & 0.604/0.413
  & 0.625/0.421
  & 111,270/13,007,654 = 0.86\% \\

FlowAdapt$^{\dagger}$
  & \underline{0.574/0.338}
  & \underline{0.631/0.389}
  & \underline{0.672/0.458}
  & \underline{0.689/0.481}
  & \underline{0.712/0.526}
  & 95,694/12,992,078 = 0.74\% \\

FlowAdapt
  & \textbf{0.599/0.354}
  & \textbf{0.650/0.414}
  & \textbf{0.698/0.492}
  & \textbf{0.715/0.521}
  & \textbf{0.738/0.555}
  & 137,393/13,033,777 = 1.05\% \\

\bottomrule
\end{tabular}
}
\vspace{-3pt}
\caption{Performance comparison across different deployment data ratios (1\%-20\%). The baseline model CoAlign is pre-trained on OPV2V and adapted to DAIR-V2X. The top two results are highlighted using \textbf{bold} and \underline{underlined} fonts respectively.}
\label{op2dair_performance}
\vspace{-5pt}
\end{table*}

\begin{figure}[t]
\centering
\includegraphics[width=1.04\linewidth]{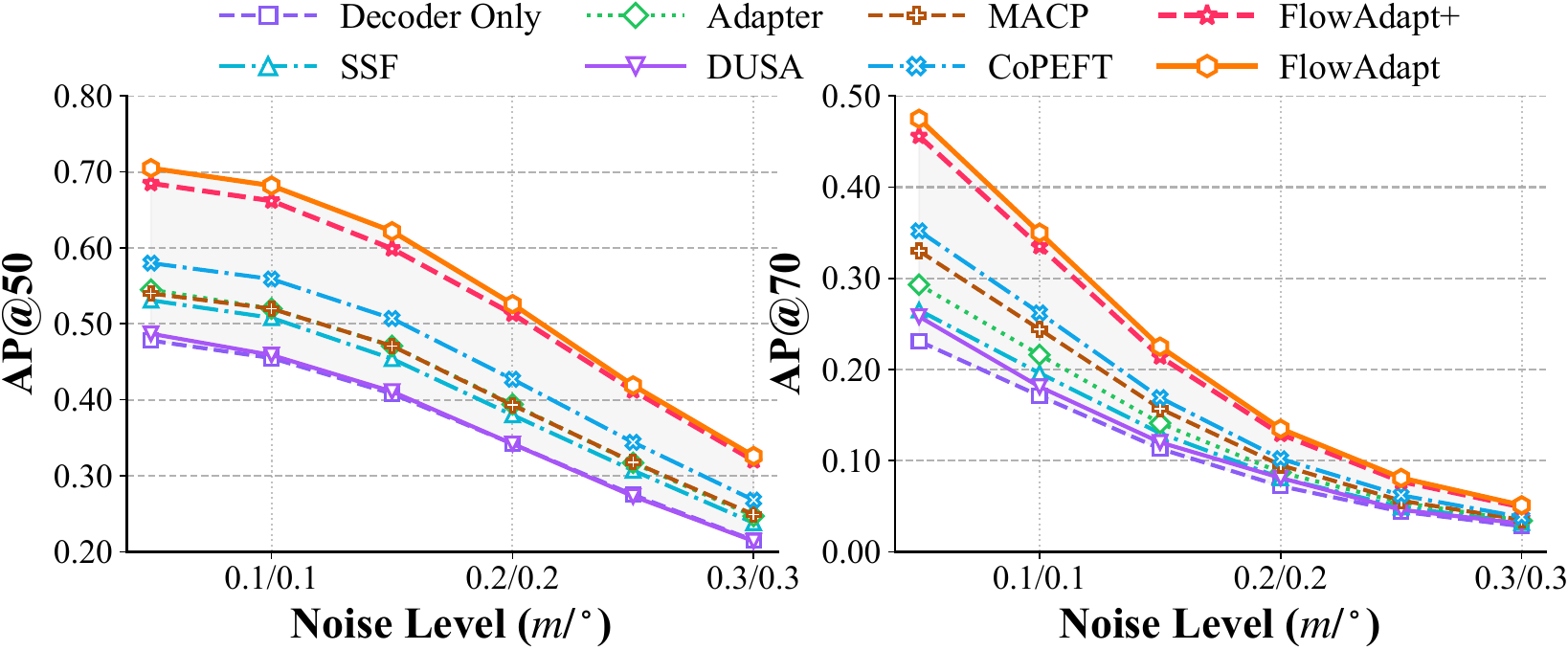}
\caption{Performance comparison under localization noise.}
\label{fig:noise_robustness}
\end{figure}

\begin{table}[t]
\centering
\small
\setlength{\tabcolsep}{8pt}
\renewcommand{\arraystretch}{0.95}
\begin{tabular}{lccc}
\toprule
Method & AttFuse & MKD-Cooper & FreeAlign \\
\cmidrule(lr){2-4}
 & \multicolumn{3}{c}{AP@50/70~$\uparrow$} \\
\midrule
None
  & 0.439/0.205 & 0.317/0.158 & 0.341/0.188 \\
From scratch
  & 0.307/0.162 & 0.467/0.314 & 0.352/0.193 \\
Adapter
  & 0.495/0.291 & 0.517/0.301 & 0.514/0.293 \\
LoRA
  & 0.401/0.220 & 0.439/0.300 & 0.437/0.301 \\
ResLoRA
  & 0.386/0.201 & 0.446/0.301 & 0.468/0.327 \\
DUSA
  & 0.475/0.317 & 0.448/0.304 & 0.470/0.314 \\
MACP
  & 0.518/0.326 & 0.525/0.337 & 0.529/0.322 \\
CoPEFT
  & 0.520/0.352 & 0.533/0.349 & 0.536/0.336 \\
FlowAdapt$^{\dagger}$
  & \underline{0.650/0.470} & \underline{0.659/0.458} & \underline{0.661/0.456} \\
FlowAdapt
  & \textbf{0.669/0.496} & \textbf{0.683/0.474} & \textbf{0.677/0.469} \\
\bottomrule
\end{tabular}
\vspace{-3pt}
\caption{Generalization across different fusion architectures.}
\label{tab:generalization_fusion}
\vspace{-8pt}
\end{table}

\begin{table}[t]
\centering
\small
\setlength{\tabcolsep}{16pt}
\renewcommand{\arraystretch}{0.95}
\begin{tabular}{lcc}
\toprule
Method & V2V4Real & V2XSet \\
\cmidrule(lr){2-3}
 & \multicolumn{2}{c}{AP@50/70~$\uparrow$} \\
\midrule
None
  & 0.461/0.217 & 0.917/0.839 \\
From scratch
  & 0.458/0.213 & 0.875/0.708 \\
Adapter
  & 0.559/0.260 & 0.932/0.845 \\
LoRA
  & 0.538/0.270 & 0.880/0.717 \\
ResLoRA
  & 0.535/0.260 & 0.891/0.756 \\
DUSA
  & 0.533/0.264 & 0.886/0.847 \\
MACP
  & 0.578/0.284 & 0.933/0.852 \\
CoPEFT
  & 0.548/0.270 & 0.931/0.851 \\
FlowAdapt$^{\dagger}$
  & \underline{0.618/0.393} & \underline{0.939/0.874} \\
FlowAdapt
  & \textbf{0.626/0.413} & \textbf{0.942/0.878} \\
\bottomrule
\end{tabular}
\vspace{-3pt}
\caption{Generalization across different target domains.}
\label{tab:generalization_domain}
\vspace{-8pt}
\end{table}

\subsection{Quantitative Evaluation}

\subsubsection{Adaptation from OPV2V to DAIR-V2X.}
Table~\ref{op2dair_performance} reports adaptation across labelling ratios from 1\% to 20\%. Without adaptation the source model transfers poorly, and training from scratch fares worse on the few labelled frames. Among PEFT baselines, SSF~\cite{ssf} and Adapter stay light but plateau early, while LoRA and ResLoRA spend far more parameters, without overtaking the lighter two in the low-data regime. FlowAdapt leads at every ratio in both AP@50 and AP@70 with about 1\% trainable parameters, exceeding CoPEFT, the strongest baseline under this setting, by 11.1 and 10.8 points at 10\%. Its lead over CoPEFT widens as labels grow scarce, the regime this work targets. The unsupervised DUSA instead draws on the whole target pool without labels and updates all parameters, trading supervision for cost differently, while the lighter FlowAdapt$^{\dagger}$ at 0.74\% parameters retains most of FlowAdapt's advantage throughout.

Figure~\ref{fig:noise_robustness} probes adaptation under localization noise, perturbing each non-ego pose on DAIR-V2X\@. AP@50 and AP@70 fall for every method as misalignment grows, yet FlowAdapt stays ahead of CoPEFT across the whole range, and FlowAdapt$^{\dagger}$ follows it closely at every level.

\subsubsection{Generalization across Architectures and Domains.}
With the collaborative detector source-trained under AttFuse, MKD-Cooper~\cite{mkdcooper} or FreeAlign~\cite{freealign} rather than CoAlign, FlowAdapt leads CoPEFT by 14 to 15 points in AP@50 and 12 to 14 in AP@70 on all three, so the adaptation is not tied to the tested fusion designs. Across target domains the lead instead varies with the shift. On V2V4Real, a real vehicle-to-vehicle target, FlowAdapt gains 7.8 and 14.3 points over CoPEFT, whereas on the simulated V2XSet, which shares the CARLA source of OPV2V and leaves every method little room, the gain falls to 1.1 and 2.7. The larger gains track the real targets, not the simulated one.

\begin{figure}[t]
\centering
\includegraphics[width=1.0\linewidth]{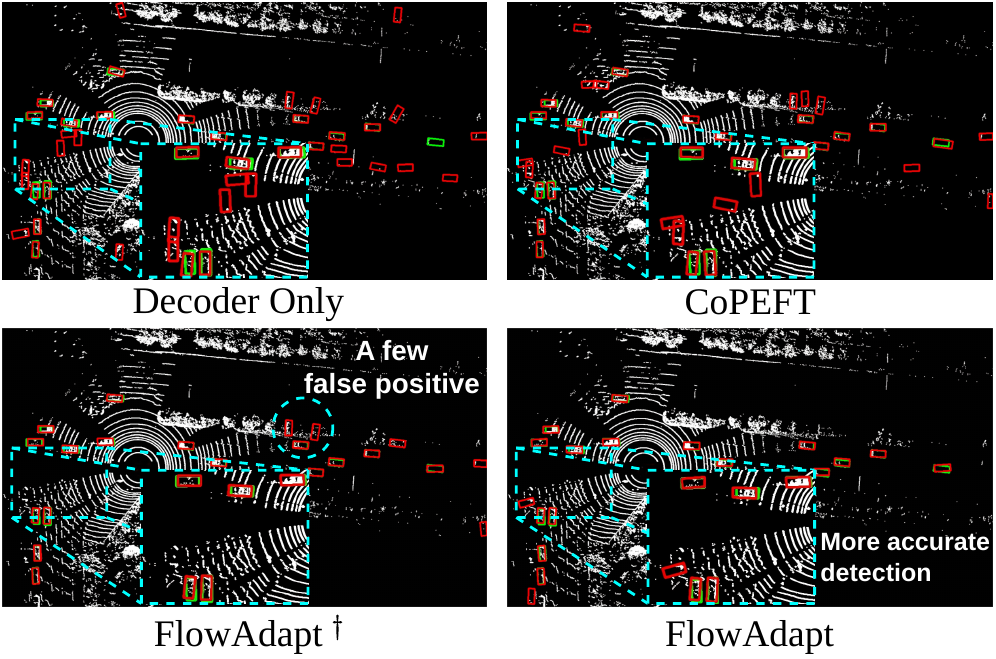}
\caption{Comparison of detection results on DAIR-V2X adapted from OPV2V. Green and red 3D bounding boxes represent ground truth and predictions, respectively.}
\label{fig:visual}
\end{figure}

\begin{figure}[t]
\centering
\includegraphics[width=1.0\linewidth]{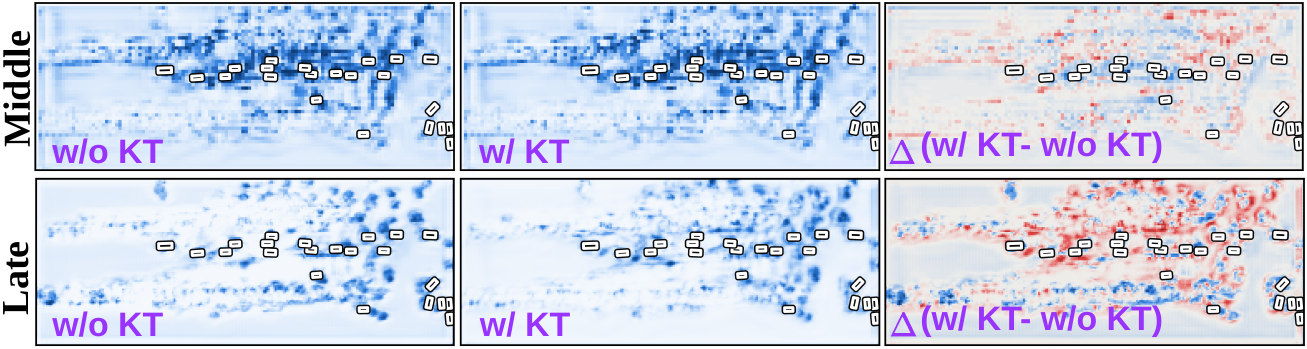}
\caption{Linear-probe foreground confidence at the middle and late stages, without and with cross-stage transfer.  Blue is higher confidence and red marks where transfer raises it.}
\vspace{-5pt}
\label{fig:probe_heatmap}
\end{figure}

\begin{figure}[t]
\centering
\includegraphics[width=1.0\linewidth]{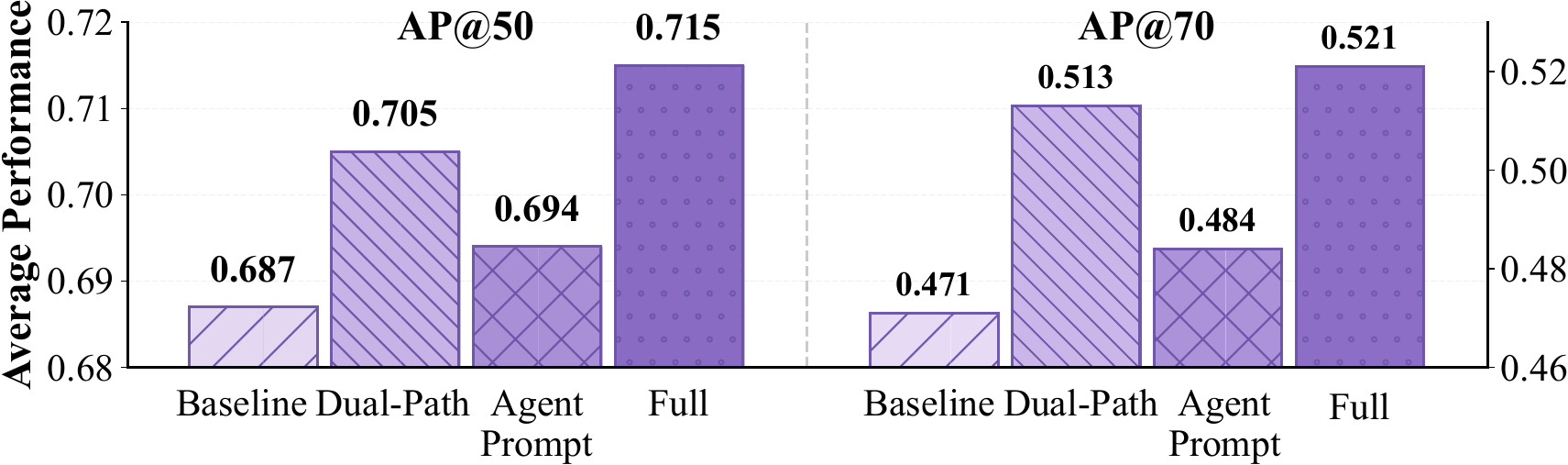}
\caption{Ablation study on key components.}
\label{fig:key_design}
\end{figure}

\begin{table}[t]
\centering
\small
\setlength{\tabcolsep}{6pt}
\renewcommand{\arraystretch}{0.5}
\begin{tabular}{cccccccc}
\toprule
\multirow{2}{*}{WGS} & \multicolumn{5}{c}{KTPro} & \multirow{2}{*}{AP@50} & \multirow{2}{*}{AP@70} \\
\cmidrule(lr){2-6}
 & E & M & L & M+KT & L+KT & & \\
\midrule
- & \multicolumn{5}{c}{-} & 0.483 & 0.256 \\
\checkmark & \multicolumn{5}{c}{-} & 0.495 & 0.285 \\
- & \multicolumn{5}{c}{\checkmark} & 0.600 & 0.431 \\
\midrule
\checkmark & \checkmark & - & - & - & - & 0.685 & 0.485 \\
\checkmark & \checkmark & \checkmark & - & - & - & 0.697 & 0.492 \\
\checkmark & \checkmark & \checkmark & \checkmark & - & - & 0.700 & 0.501 \\
\checkmark & \checkmark & \checkmark & \checkmark & \checkmark & - & 0.706 & 0.512 \\
\checkmark & \checkmark & \checkmark & \checkmark & \checkmark & \checkmark & \textbf{0.715} & \textbf{0.521} \\
\bottomrule
\end{tabular}
\vspace{-3pt}
\caption{Ablation study on FlowAdapt. E/M/L: early/middle/late stage adaptation. M/L$+$KT: knowledge transfer from the early stage into the middle or late stage.}
\label{tab:ablation}
\end{table}

\begin{table}[!t]
\centering
\vspace{-3pt}
\small
\setlength{\tabcolsep}{3.5pt}
\renewcommand{\arraystretch}{0.85}
\begin{tabular}{lcccc}
\toprule
\multirow{2}{*}{\hspace{-5pt}Method} & \multicolumn{4}{c}{Sampling Strategy (AP@50/70)} \\
\cmidrule(lr){2-5}
& Random & Uniform & K-means & WGS \\
\midrule
\hspace{-6pt}Decoder only & 0.485/0.263 & 0.486/0.265 & 0.483/0.266 & 0.495/0.285 \\
\hspace{-6pt}CoPEFT       & 0.619/0.406 & 0.624/0.412 & 0.631/0.414 & 0.643/0.428 \\
\hspace{-6pt}FlowAdapt$^{\dagger}$  & 0.666/0.473 & 0.672/0.471 & 0.675/0.471 & 0.689/0.481 \\
\hspace{-6pt}FlowAdapt    & 0.674/0.482 & 0.683/0.487 & 0.691/0.498 & \textbf{0.715}/\textbf{0.521} \\
\bottomrule
\end{tabular}
\caption{Performance comparison across sampling strategies.}
\vspace{-10pt}
\label{tab:comparison}
\end{table}

\subsection{Qualitative Evaluation}
Figure~\ref{fig:visual} compares detections on one target scene. The decoder-only baseline both misses objects and fires on empty ground, and CoPEFT reduces the false positives while still misplacing boxes where the geometry is ambiguous. FlowAdapt$^{\dagger}$ removes most of those errors, a few false positives remaining in the marked region, and FlowAdapt clears the region and keeps the boxes tight to the ground truth.

Figure~\ref{fig:probe_heatmap} shows that without transfer the late-stage probe scores the foreground weakly, its confidence scattered off the objects and onto empty road, while the middle stage separates the two more sharply. Transfer raises both, and the difference map places the gain on the foreground, red concentrating on the boxes and their immediate surroundings rather than the background. The gain is largest at the late stage, where decodability starts lowest and the route KTPro opens ends.


\subsection{Ablation Study}

\subsubsection{Component Contributions.}
Table~\ref{tab:ablation} decomposes the 23.2-point AP@50 gain FlowAdapt brings over the frozen baseline. Drawing the 10\% by WGS rather than from the leading contiguous frames accounts for 1.2 points, from 0.483 to 0.495. The early adapter then takes the largest step, to 0.685, the remaining adapters bring it to 0.700, and the cross-stage routes reach 0.715 and 0.521 in AP@50 and AP@70. These figures follow the order of addition and read as marginal contributions along one path. Dropping the routes while keeping the adapters costs 1.5 and 3.6 points, so the gate they supply is not something the deeper adapters recover alone. The same substitution is worth 11.5 points on the full model against 1.2 on the frozen one, so the two axes do not simply add.

\subsubsection{WGS and Its Synergy with KTPro.}
Table~\ref{tab:comparison} fixes the 10\% budget and varies only how the frames are chosen, the random and uniform baselines each averaged over ten runs. WGS leads every alternative, gaining 4.1 points in AP@50 over random for FlowAdapt against 1.0 for the decoder-only model. It also stays ahead of K-means run on the same configurations, which clusters by an average squared distance where WGS minimizes the largest one. The margin widens monotonically across the four models, from 1.0 to 4.1 points.

\subsubsection{Effectiveness of Adapter Design Choices.}
Figure~\ref{fig:key_design} starts from a baseline with neither design and adds each in turn. The dual path carries most of the effect, its spatial and channel branches together lifting AP@70 from 0.471 to 0.513. The collaborative prompt then adds a further gain to 0.521, its contribution measured with the dual path already in place.

\section{Conclusion}
We revisit PEFT adaptation for collaborative perception from an optimal transport perspective and present FlowAdapt, a parameter-efficient framework coupling Wasserstein Greedy Sampling and Progressive Knowledge Transfer. Two bottlenecks have been studied apart rather than together, the redundancy among the configurations a target sequence presents, which WGS addresses by covering them under a transport criterion, and the lower accuracy of a linear probe at the deeper stages of a frozen backbone, which KTPro addresses by gating the corrections proposed there. Training 1\% of the parameters on 10\% of the target labels, FlowAdapt substantially improves adaptation accuracy and keeps its lead across target domains, fusion architectures and localization noise.

\bibliography{aaai2027}


\end{document}